\documentclass[10pt,twocolumn,letterpaper]{article}

\usepackage{cvpr}
\usepackage{times}
\usepackage{epsfig}
\usepackage{graphicx}
\usepackage{amsmath}
\usepackage{textcomp}
\usepackage{amssymb}

\usepackage[pagebackref=true,breaklinks=true,letterpaper=true,colorlinks,bookmarks=false]{hyperref}

\cvprfinalcopy 

\def\cvprPaperID{41} 

\ifcvprfinal\fi
\begin{document}

\title{Reducing catastrophic forgetting with learning on synthetic data}

\author{Wojciech Masarczyk\\
Institute of Theoretical and Applied Informatics,\\
Polish Academy of Sciences\\
Tooploox\\
{\tt\small wojciech.masarczyk@gmail.com}
\and
Ivona Tautkute\\
Polish-Japanese Academy of Information Technology \\
Tooploox\\
{\tt\small ivona.tautkute@tooploox.com}
}





\maketitle

\begin{abstract}
Catastrophic forgetting is a problem caused by neural networks' inability to learn data in sequence. After learning two tasks in sequence, performance on the first one drops significantly. This is a serious disadvantage that prevents many deep learning applications to real-life problems where not all object classes are known beforehand; or change in data requires adjustments to the model. To reduce this problem we investigate the use of synthetic data, namely we answer a question: Is it possible to generate such data synthetically which learned in sequence does not result in catastrophic forgetting? We propose a method to generate such data in two-step optimisation process via meta-gradients. Our experimental results on Split-MNIST dataset show that training a model on such synthetic data in sequence does not result in catastrophic forgetting. We also show that our method of generating data is robust to different learning scenarios.

\end{abstract}

\section{Introduction}

\begin{figure*}[!t]
    \centering
    \includegraphics[width=\textwidth]{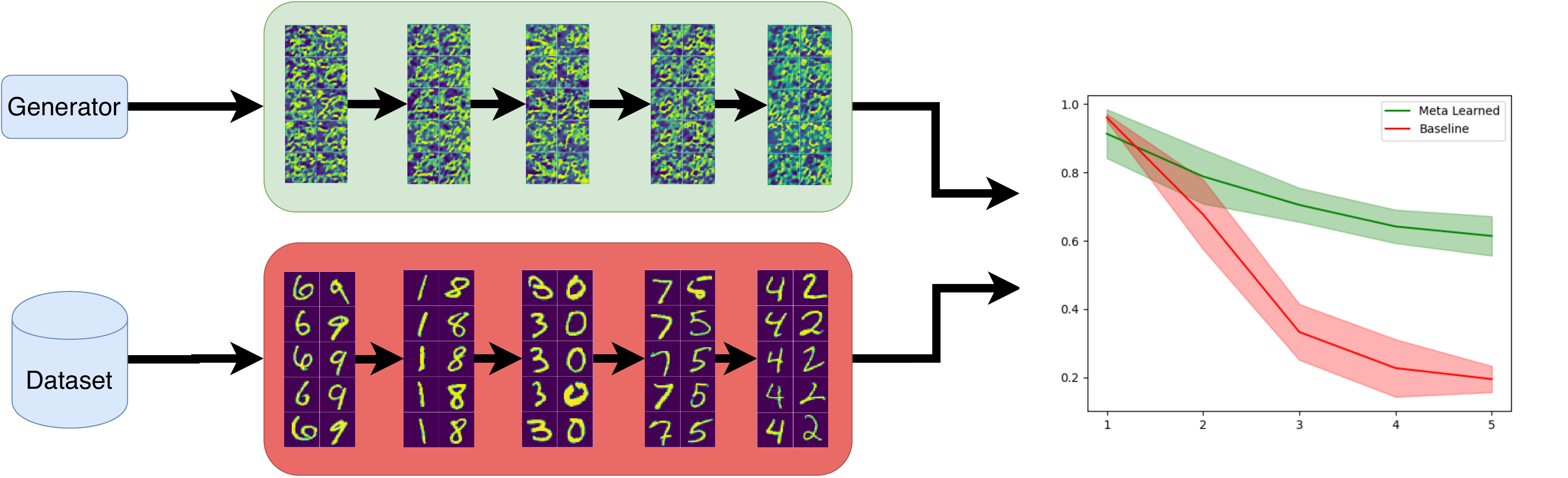}
    \caption{Synthetic data created from generator is divided into five tasks according to classes and learner (green) learns tasks sequentially. The same procedure is applied to learner with real data (red). The right plot shows that accuracy at the end of each task does not decrease on learned data in contrast to real data where it deteriorates sharply.}
    \label{fig:method}
\end{figure*}{}

Deep learning methods have succeeded in many different domains such as: scene understanding, image generation, natural language processing \cite{devlin18bert, tautkute18deepstyle, stokowiec17shallowread, oleszkiewicz18siamese}. While deep learning methods differ in architecture choice, objective function or optimization strategy, they all assume that the training data is independent and identically distributed (i.i.d). 
Methods built on this assumption are effective for fixed environments with stationary data distributions -- where tasks to be solved do not change over time or classes present in the dataset are known from the beginning. However, in most real-life scenarios this assumption is violated and there is a need for methods that are able to handle such cases.
Among many examples of such scenarios, a few can be highlighted: new object class is introduced, however the dataset used to train the baseline model is no longer available; the data characteristics seem to change seasonally and model needs to change its predictions accordingly to these trends.
Continual learning \cite{ring94thesis} is a paradigm where data is presented sequentially to the algorithm without the ability to manipulate this sequence. Additionally, there is no assumption about the structure of the sequence. A successful continual learning algorithm needs to be able to learn a
growing number of tasks, be resistant to catastrophic forgetting \cite{mccloskey89catastrophic} and be able to adapt do distribution shifts. The memory and computational requirements of such algorithm should scale reasonably with the incoming data.

Although the problem of continual learning is known for many years \cite{ring94thesis, mccloskey89catastrophic}, only recently has the field gained significant traction and many interesting ideas have been proposed.
Most of continual learning contributions can be divided into three categories \cite{lomonaco18thesis, parisi19review}: optimization, architecture and rehersal. Methods based on optimization modifications usually add additional regularization terms to objective function to dampen catastrophic forgetting \cite{kirkpatrick16ewc, li16lwf}. Second category gathers methods that propose various architectural modifications e.g. Progressive Net \cite{rusu2016progressive} where increasing capacity is obtained by initialising new network for each task. The last category -- rehersal based methods -- consists of methods that assume life-long presence of a subset of historical data that can be re-used to retain knowledge about past tasks \cite{lopez17gem, hayes18exstream}.

This work proposes a new data-driven path that is orthogonal to existing approaches. Specifically, we would like to explore the possibility of creating input data artificially in a coordinated manner in such a way that it reduces the catastrophic forgetting phenomena. We achieve this by combining two separate neural networks connected by two-step optimisation.
We use generative model to create synthetic dataset and form a sequence of tasks to evaluate learner model in continual learning scenario. The sequence of synthetic tasks is used to train the learner network. Then, the learner network is evaluated on real data. The loss obtained on real data is used to tune the parameters of the generative network. In the following step, the learning network is replaced with a new one.

Differently from existing approaches, our method is independent of training method and task and it can be easily incorporated to above-mentioned strategies providing additional gains.

\section{Related Work}
One line of research for continual learning focuses on optimization process. It draws inspiration from the biological phenomena known as synaptic plasticity \cite{cichon15synaptic}. It assumes that weights (connections) that are important for particular task become less plastic in order to retain the desired performance on previous tasks. An example of such approach is Elastic Weight Consolidation (EWC) \cite{kirkpatrick16ewc}, where regularisation term based on Fisher Information matrix is used to slow down the change of important weights. However accumulation of these constrains prevents network from learning longer sequences of tasks.
Another optimization based method is Learning without Forgetting (LwF)~\cite{li16lwf}. It tries to retain the knowledge of previous tasks by optimizing linear combination of current task loss and knowledge distillation loss. LwF is conceptually simple method that benefits from knowledge distillation phenomenon \cite{hinton15distillation}. The downside of such approach is that applying LwF requires additional memory and computation resources for each optimization step.

Methods based on architectural modifications allow to dynamically expand/shrink networks, select sub-networks, freeze weights or create additional networks to preserve knowledge. Authors of \cite{rusu2016progressive} propose algorithm that for each new task creates a separate network (a column) that is trained to solve particular task. Additionally, connections between previous columns and the current column are learned to enable forward transfer of knowledge. This algorithm avoids catastrophic forgetting completely and enables effective transfer learning. However the computational cost of this approach is prohibitive for longer sequences of tasks.
Other methods \cite{lee17dynamic, wang19growing} address the problem of computational cost by expanding single layers/neurons instead of whole networks, however these methods has less capacity to solve upcoming tasks.
Different approaches that modify architectures are based on selecting sub-networks used for solving current task in such a way that only a fraction of network's parameters relevant to current task is changed \cite{mallya18piggyback, mallya17packnet, golkar19continualviapruning}. The challenge here is to 
balance the number of frozen and active weights in such way that network is still able to learn new tasks and preserve current knowledge.

Rehearsal methods are based on the concept of memory replay. It is assumed that subset of previously processed data is stored in memory bank and interleaved with upcoming data in such a way that neural network learns to solve current task in addition to preserving current knowledge \cite{lopez17gem, wang18datasetdistillation, hayes18exstream}.
A naive rehearsal method would be to save random data samples that were 
present during training. However such approach is inefficient, since samples are not equally informative, hence the challenge of rehearsal methods is to choose the most representative samples for a given dataset, such that minimum storage is occupied. In \cite{wang18datasetdistillation}, authors apply method of dataset distillation based on meta-gradient optimization to reduce the size of memory bank. It is possible to represent whole class of examples just by storing one carefully-optimized example. Unfortunately, applying this meta-optimization method is computationally exhaustive. The biggest downside of using rehearsal based methods is the need to store the actual data which in some cases can violate data privacy rules or can be computationally prohibitive. To mitigate this issue solution based on Generative Networks was proposed~\cite{wu18mergan, shin17generativereplay}. Namely, they use dual model architecture composed of learner network and generative network. Role of the generative network is to model data previously experienced by the learner network. Data sampled from the generator network is used as a rehersal data for learner network to reduce the effect of catastrophic forgetting.

Our method is also dual architecture model based on generative network, however the aim of generative network is radically different. In contrast to authors \cite{wu18mergan, shin17generativereplay} we do not aim to capture the statistics of real data, instead we try to generate entirely synthetic data such that when learner does learn on a sequence of such data it does not suffer from catastrophic forgetting.

\section{Method}
\label{section:method}
\begin{figure}[!t]
    \centering
    \includegraphics[width=\columnwidth]{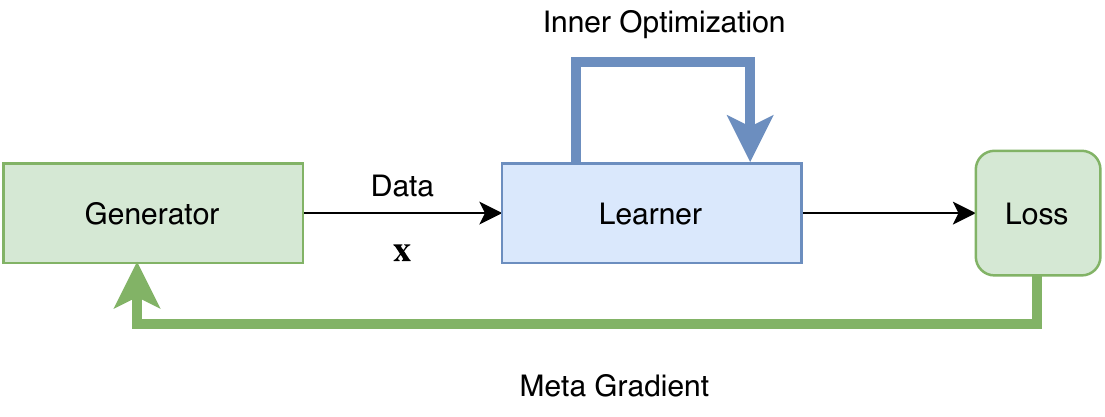}
    \caption{Synthetic data from generator is passed to learner where the inner optimization is performed and meta-loss is backpropagated to $\mathcal{G}$.}
    \label{fig:meta_train_scheme}
\end{figure}

The main idea of our approach is to generate data samples such that network trained on them in sequence would not suffer from catastrophic forgetting. One of many ways to generate artificial data is to use meta-optimization strategy introduced in \cite{maclaurin15reversible}. It is shown that by applying meta-learning it is possible to use gradient optimization both to hyperparameters and to input data. However, this approach is limited to small problems, since each data point must be optimised separately. To overcome this bottleneck, authors of Generative Teaching Networks (GTNs) \cite{such2020generative} use generative network to create artificial data samples instead of directly optimizing the data input. 
We adopt similar approach in our method, namely, we use generative network -- green rectangle ''Generator'' in Fig. \ref{fig:meta_train_scheme} -- to produce synthetic data from noise vectors sampled from a random distribution. Next, we split the data into separate tasks according to classes and form a continual learning task for the learner network -- blue rectangle in Fig. \ref{fig:meta_train_scheme}.
Learner network after completing whole sequence of tasks in evaluated on 
real training data. The loss from real data classification after learning all tasks in sequence is then backpropagated to generator network to tune the parameters as shown in Fig. \ref{fig:meta_train_scheme}.

Our approach is similar to one proposed in work \cite{khurram19oml}. Using two step meta-learning optimization they try to learn best representation of input data such that the model learned in with standard optimization does not suffer from catastrophic forgetting.

Differently from \cite{such2020generative}, we do not use curriculum based learning as our goal is to have a realistic continual learning scenario where the order of data sequence is not known beforehand. To ensure that the Generator network does not generate data suitable for particular sequence of tasks at each meta-optimization we shuffle order of tasks. Precisely, at each step we generate $p$ samples for each class and then randomly create a sequence of binary classification tasks with particular data.

{}
 
Precisely, let $\mathcal{G}$ be a generative neural network, $\mathcal{S}$ a standard convolutional network for classification, $\mathbf{t} = (t_1, t_2, \ldots t_n)$ a sequence of tasks, where each tasks is binary classification task and classes in each task form mutually disjoint sets.

The inner training loop consists of sequence of tasks, where generated samples from previous tasks are not replayed once the task is finished. To achieve this, the sequence of tasks $\mathbf{t} = (t_1, t_2, \ldots t_n)$ must be defined a priori and samples generated by network $\mathcal{G}$ are conditioned on the information of particular task.  For each task $t_i$  the network $\mathcal{G}$ generates two batches of samples $\mathbf{x} = \mathcal{G}(\mathbf{z}, \mathbf{y_{i_j}})$ for $j=1,2$, where $\mathbf{z}$ is a batch of noise vectors sampled from Normal distribution and $\mathbf{y_{i_j}}$ is a class indicator for task $\mathbf{t_i}$.
Note that generator networks has access to class indicators since we aim to learn in continual learning scenario only the learner network.

Neural network $\mathcal{S}$ learns sequentially on following tasks using standard SGD optimizer with learning rate and momentum optimized through meta-gradients. At the end of the sequence $\mathbf{t}$ network $\mathcal{S}$ is evaluated on real dataset ($\mathbf{x_r}, \mathbf{y_r}$) obtaining meta-loss as shown in Fig. \ref{fig:meta_train_scheme}. This meta-loss is backpropagated through all training inner-loops of model $\mathcal{S}$ to optimize network $\mathcal{G}$.
Parameters $\theta$ of network $\mathcal{G}$ are updated according to the equation:
\begin{equation}
    \theta = \theta - \eta \nabla_\theta \mathcal{L} (\mathcal{S}(\mathbf{x_r}; \mathbf{w_m}), \mathbf{y_r}),
    \label{eq:g_update}
\end{equation}
where $\mathbf{w_m}$ are parameters of the network $\mathcal{S}$ after $m$ optimization steps, $\eta$ is fixed learning rate, $\mathcal{L}$ is a cross entropy loss function, $\mathbf{x_r}, \mathbf{y_r}$ are real data samples and labels respectively.

\section{Experiments}
To test our hypothesis we use popular continual learning benchmark Split-MNIST \cite{lee17moment, srivastava13compete}. In first experiment, we use 5-fold split with two classes for each task to create a moderately difficult sequence of tasks. Network $\mathcal{G}$ generates 250 samples per each class. During inner optimisation learner network is optimized on batch size formed with 40 generated images (20 samples per class drawn randomly from the pool of 250 samples per class). We train the learner network on each task for 5 inner steps with batch size 40. Once the task is over, samples from this task are not shown to the network to the end of training. At test time, after learning on each task the network is evaluated on part of a test set composed of classes seen in previous taks. Both networks are simple convolutional neural networks with two convolutional layers with addition of one and two fully connected layers for classification and generative network respectively. Each layer is followed by a batch normalisation layer.

\begin{figure}[!b]
    \centering
    \includegraphics[width=\columnwidth]{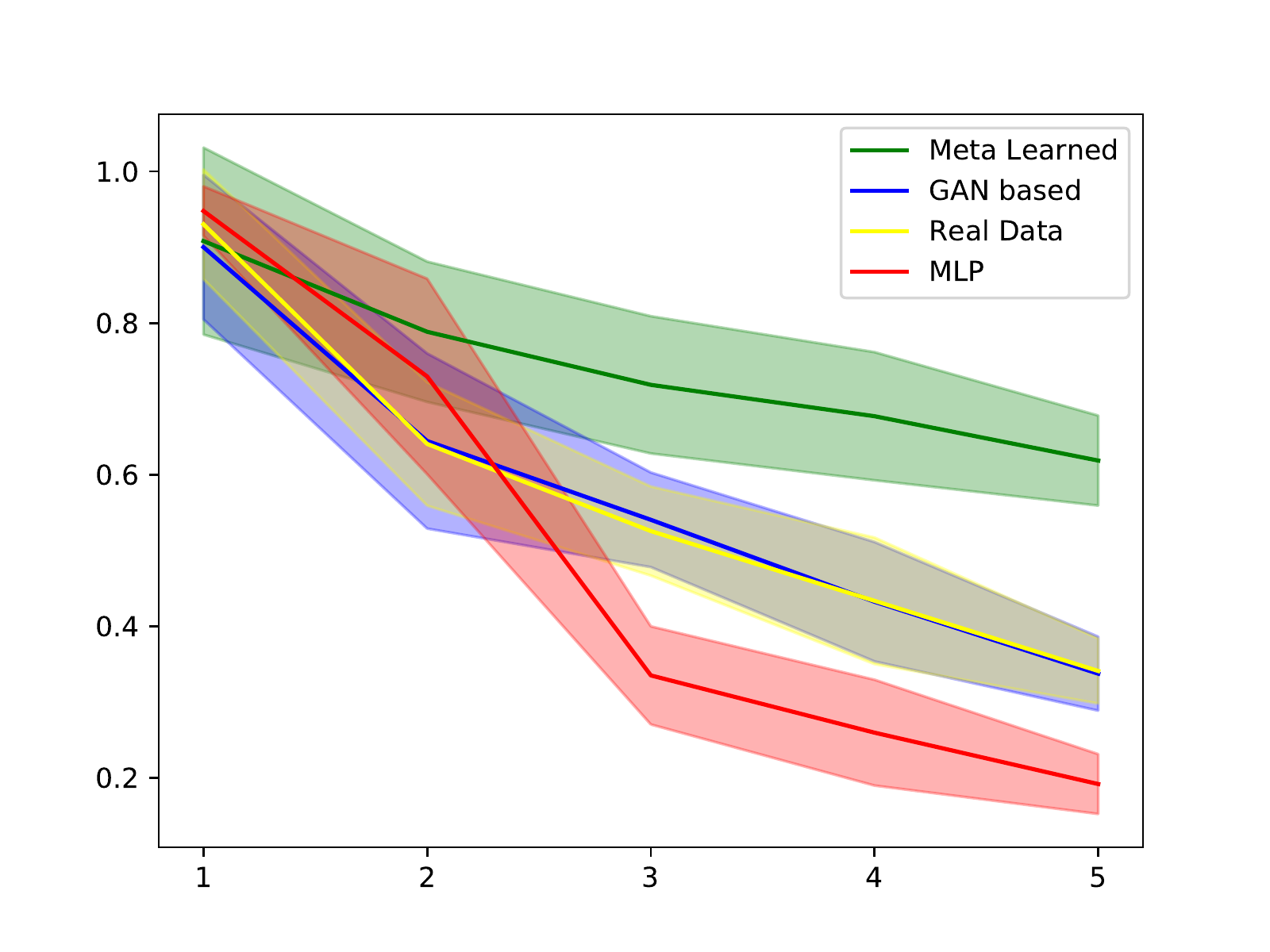}
    \caption{Overall accuracy measured on test data subset. After learning each task, test data subset is  made of samples only
    from classes seen during recent and previous tasks.}
    \label{fig:real_vs_learned}
\end{figure}

As a baseline to compare with, we use simple fully connected network proposed in \cite{kemker17mesuring} ('MLP' -- red -- in Fig. \ref{fig:real_vs_learned}). To further investigate the impact of generated data we use the same network architectures and optimizer settings with learning rate and momentum optimized with by a meta learning process as described in Section \ref{section:method} but for optimizing the learner network we use real data ('Real Data' -- yellow -- in Fig. \ref{fig:real_vs_learned}).
We also compare our results with GAN-based data samples. In this scenario we follow the setting of 'Real Data' scenario except for the source of data. We use Conditional-GAN \cite{mirza14cgan} to model the original data distribution and then sample 250 samples per each class ('GAN based' -- blue -- in Fig. \ref{fig:real_vs_learned}).

We implement experiments in PyTorch library, which is well suited for computing
higher-order gradients \cite{grefenstette2019generalized}.
\begin{figure}
    \centering
    \includegraphics[width=1\columnwidth]{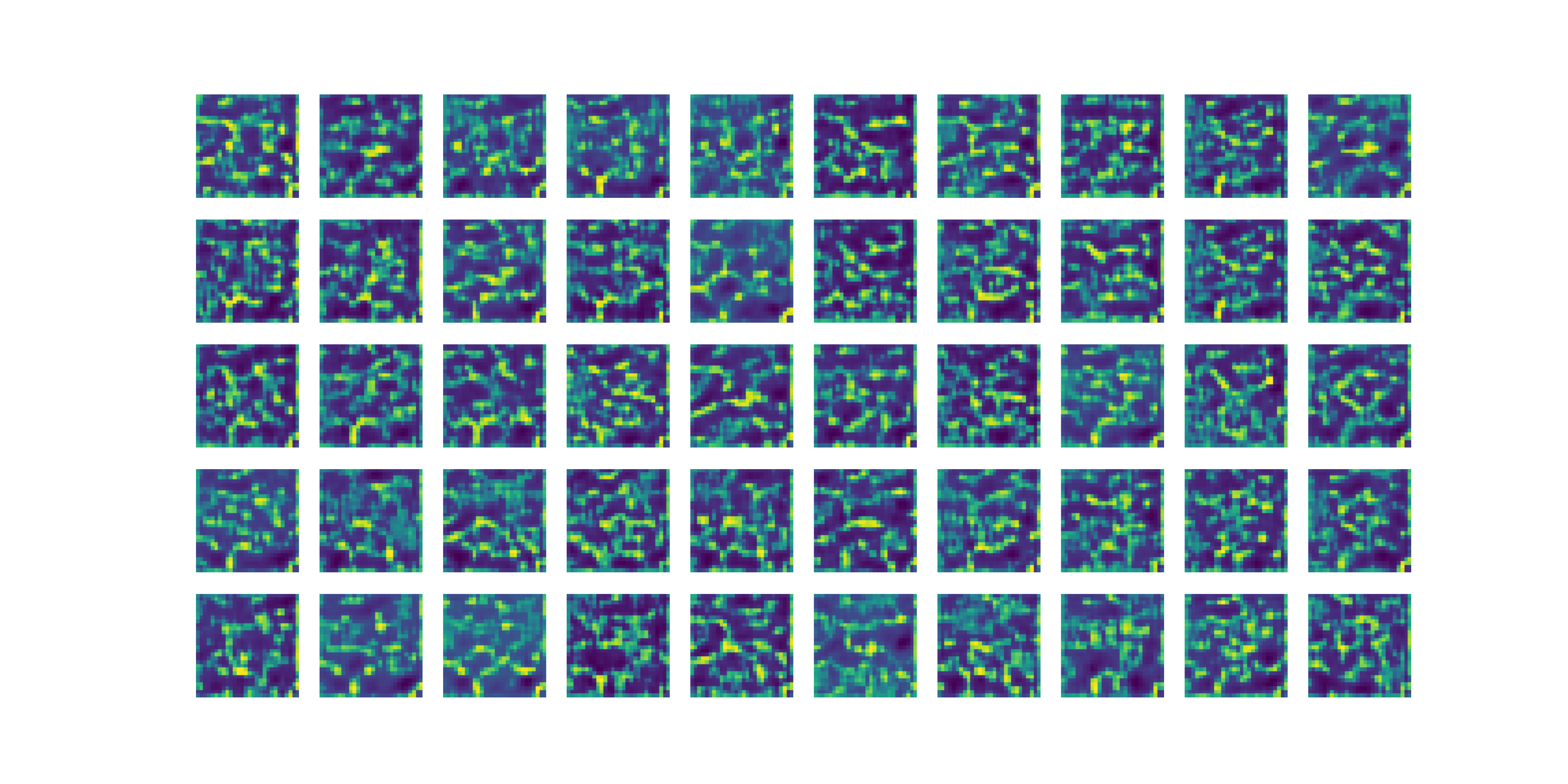}
    \caption{Samples generated by network $\mathcal{G}$ at the end of meta-optimisation. Starting from zero (leftmost), each sample to the right represents the following class.}
    \label{fig:samples}
\end{figure}{}

{}

\begin{figure}
    \centering
    \includegraphics[width=\columnwidth]{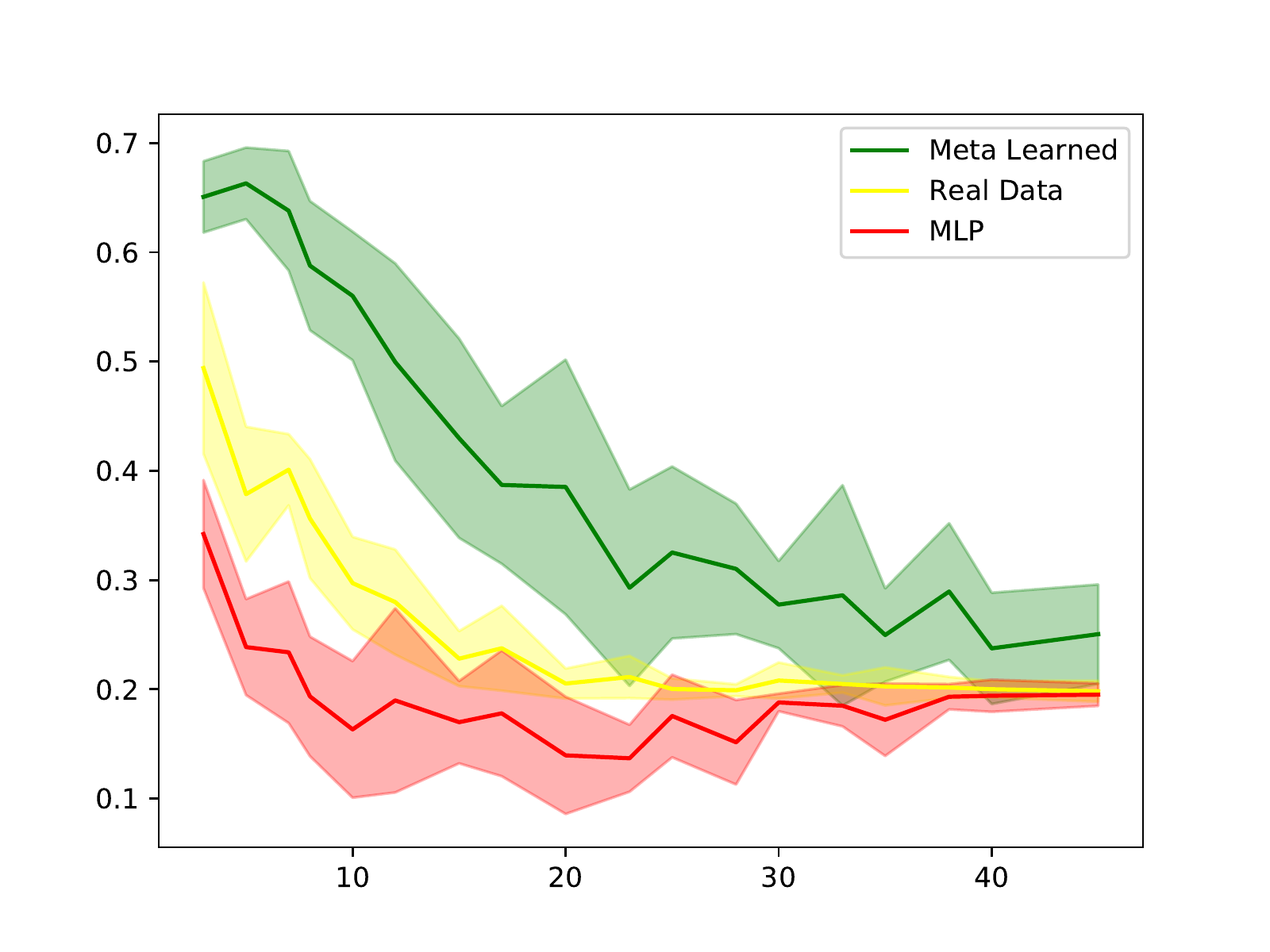}
    \caption{Overall accuracy measured on test set after learning network
    $\mathcal{S}$ with synthetic data for $x$ inner steps on each task. }
    \label{fig:inner_steps}
\end{figure}{}
\textbf{Results} -- obtained results support our hypothesis, that it is possible to generate synthetic data such that, even if networks learns this data in sequence (one time per sample), the learning process does not result in castastrophic forgetting. 

Figure \ref{fig:real_vs_learned} shows how learning on synthetic data in sequence results in less catastrophic forgetting compared to learning on a sequence of real data samples. Note that additional performance could be gained with careful hyperparameter tuning, however we did not want to compete for best performance and rather show the potential of this approach.
Higher accuracy of 'Real data' scenario over 'MLP' can be attributed to the effectiveness of optimised learning rate and momentum parameters, however the main advantage comes from using meta learned data samples. Results obtained with data generated with GAN are almost identical to ones obtained with real data. This result is expected as the data modeled by a GAN resembles original data closely.

An example batch of generated samples is shown in Figure \ref{fig:samples}. The samples are ordered according to classes (starting from 0). In contrast to \cite{such2020generative} the data samples are abstract blobs, rather than interpretable images. We verify experimentally that the reason for the lack of structure in generated samples is the lack of curriculum learning in our scenario. We skip it intentionally to provide more realistic continual learning scenario for the learner network.

Fig. \ref{fig:inner_steps} shows the impact of change of learning scenario of network $\mathcal{S}$ \textit{after} network $\mathcal{G}$ is trained.
In this experiment data generated by a network $\mathcal{G}$ in first experiment is used. Here, we investigate how the final accuracy after learning five consecutive tasks changes with the number of inner optimization steps. Note that $\mathcal{G}$ was optimised to create samples that are robust to catastrophic forgetting with inner optimization loop of 5 steps. As we can see, in case of longer learning horizon, network learned on synthetic (green plot Fig. \ref{fig:inner_steps}) data suffers significantly less than the same network learned on real data (yellow plot Fig. \ref{fig:inner_steps}). Even though accuracy of the networks drops with increasing number of inner steps, the drop is smoother in case of synthetic data.

\section{Conclusions}
The aim of this work was to answer a question, whether it is possible to create data that would dampen the effect of catastrophic forgetting. 
Experiments show that this hypothesis is true -- it is possible to generate such samples, however usually they do not visually resemble real data. Surprisingly, even applying the method alone can result in high performing network. Additional interesting advantage of this synthetic data is the robustness to changes of inner optimisation parameters -- increasing 15-fold size of a batch and length on training still results in compelling performance.
We believe that our experiments open a new and exciting path in continual learning research. As a future work we plan to adjust current method to datasets of higher complexity and test its effectiveness in online learning scenario.
\section{Acknowledgements}
Authors would like to thank Petr Hlubuček and GoodAI for publishing the code at \url{https://github.com/GoodAI/GTN}.

{\small
\bibliographystyle{ieee_fullname}
\bibliography{final}
}

\end{document}